\documentclass[sigconf]{acmart}

\AtBeginDocument{%
  }

\setcopyright{acmlicensed}
\copyrightyear{2026}
\acmYear{2026}
\acmDOI{XXXXXXX.XXXXXXX}
\acmConference[Conference acronym 'XX]{Make sure to enter the correct
  conference title from your rights confirmation email}{June 03--05,
  2018}{Woodstock, NY}
\acmISBN{978-1-4503-XXXX-X/2026/06}

\begin{document}

\title{Decoder-based Sense Knowledge Distillation}

\author{Qitong Wang}
\email{wangq19@rpi.edu}
\affiliation{%
  \institution{Rensselaer Polytechnic Institute}
  \city{Troy}
  \state{New York}
  \country{USA}
}

\author{Mohammed J. Zaki}
\email{zaki@cs.rpi.edu}
\affiliation{%
  \institution{Rensselaer Polytechnic Institute}
  \city{Troy}
  \state{New York}
  \country{USA}
}

\author{Georgios Kollias}
\email{gkollias@us.ibm.com}
\affiliation{%
  \institution{IBM Research}
  \city{Yorktown Heights}
  \state{New York}
  \country{USA}
}
\author{Vasileios Kalantzis}
\email{vkal@ibm.com}
\affiliation{%
  \institution{IBM Research}
  \city{Yorktown Heights}
  \state{New York}
  \country{USA}
}

\renewcommand{\shortauthors}{Qitong et al.}

\begin{abstract}
Large language models (LLMs) learn contextual embeddings that capture rich semantic information, yet they often overlook structured lexical knowledge such as word senses and relationships. Prior work has shown that incorporating sense dictionaries can improve knowledge distillation for encoder models, but their application to decoder as generative models remains challenging. In this paper, we introduce Decoder-based Sense Knowledge Distillation (DSKD), a framework that integrates lexical resources into the training of decoder-style LLMs without requiring dictionary lookup at inference time. Extensive experiments on diverse benchmarks demonstrate that DSKD significantly enhances knowledge distillation performance for decoders, enabling generative models to inherit structured semantics while maintaining efficient training.
\end{abstract}

\ccsdesc[500]{Computing methodologies~Natural language generation}

\keywords{Large Language Models, Representation Learning, Knowledge Distillation, Natural Language Processing}

\received{20 February 2007}
\received[revised]{12 March 2009}
\received[accepted]{5 June 2009}

\maketitle

\section{Introduction}
\begin{figure*}[ht]
\centering
\includegraphics[width=0.65\textwidth]{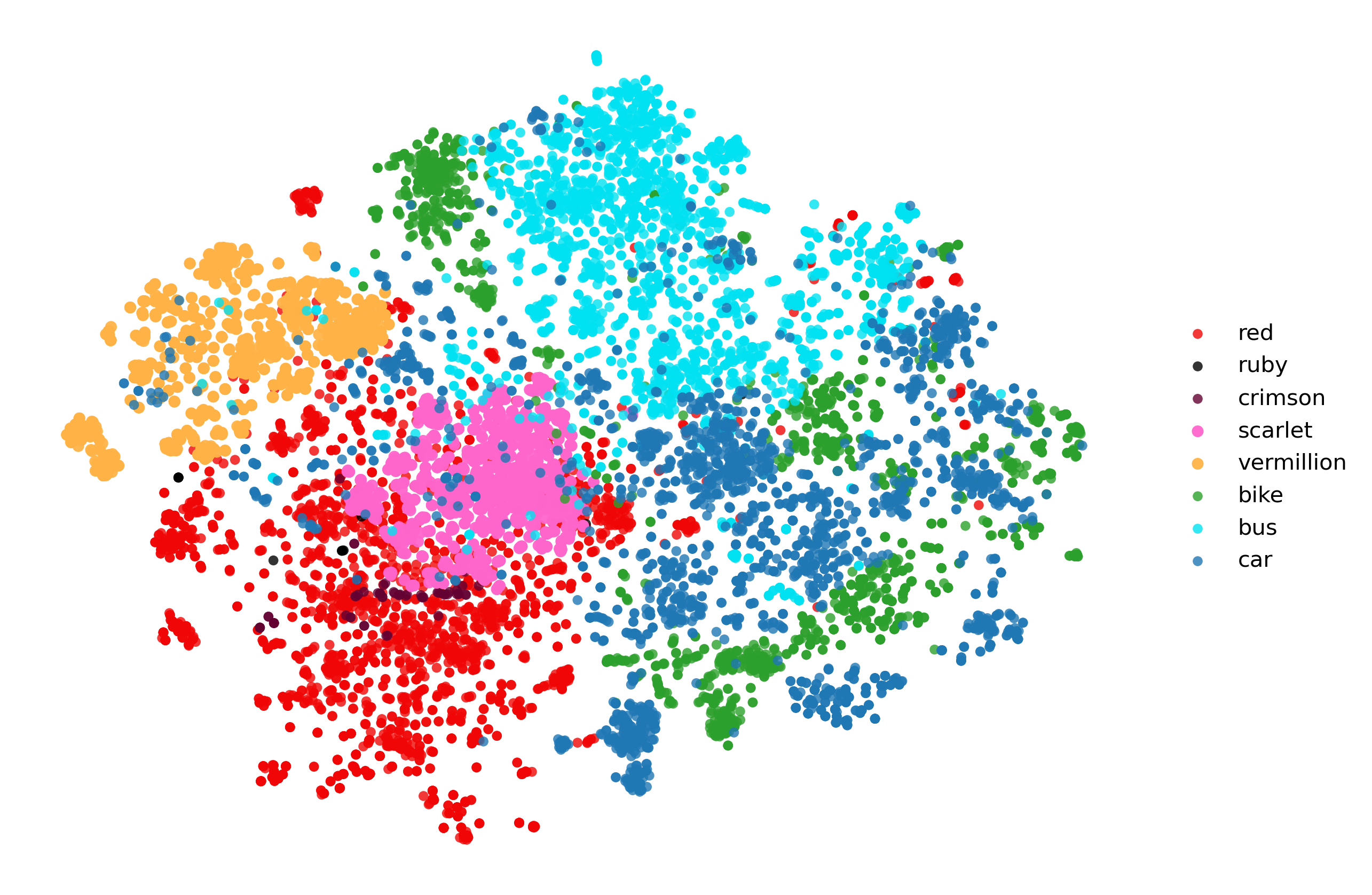}

\caption{
Distribution of selected word embeddings from Llama-3-8B-Instruct \citep{dubey2024llama} on Wikipedia dump \citep{wikipedia_dump_20240320}, with t-SNE \citep{van2008visualizing} for visualization.  Semantically related tokens form coherent clusters in the embedding space: synonyms of \emph{red} (e.g., \emph{ruby}, \emph{crimson}, \emph{vermilion}, \emph{scarlet}) appear in a compact region, while vehicle-related tokens such as \emph{bus} and \emph{car} occupy a separate neighborhood. For words that are decomposed into multiple subword tokens by the tokenizer (e.g., \emph{vermilion}, \emph{scarlet}), a composed representation is used for visualization; details are provided in Section~3.1. This structured organization suggests that contextual embeddings encode discrete semantic regularities.
}
\label{fig:red}
\end{figure*}

Language models are trained on vast collections of books, web pages, research papers, and other textual sources \citep{minaee2024large, liu2024datasets}. They learn to generate contextual embeddings, where the representation of each token depends on its surrounding context \citep{devlin2019bert, dubey2024llama,jiang2023mistral7b} . Importantly, these embeddings are not randomly distributed. Recent work on Sense Knowledge Distillation (SKD) \citep{wang2025multi}  is built on the observation that contextual embeddings produced by large language models, while continuous and highly variable, tend to form structured clusters that correspond to a limited number of semantic senses for each token. To exploit this property, SKD first constructs a sense dictionary by collecting contextual embeddings from the final hidden layer of a pretrained teacher model over a large corpus. For each token in the vocabulary, its contextual embeddings are grouped using a clustering algorithm, and the resulting cluster centroids are treated as discrete sense embeddings representing the different meanings of that token. Given this sense dictionary, SKD formulates knowledge distillation as a sense prediction problem: for each token occurrence, the teacher model identifies the most compatible sense embedding, and the student model is trained to reproduce this selection via a classification-style objective, thereby learning to align its representations with the teacher’s sense-level structure.

Nevertheless, SKD exhibits two notable limitations: (1) its adaptation to decoder architectures requires an additional LLM2Vec wrapper \citep{behnamghader2024llm2vec} that converts the decoder into an encoder, and (2) it remains confined to classification tasks, as it lacks the capacity for free-form token generation. To address these limitations, we explore the applicability of the sense dictionary to decoder models. Figure~\ref{fig:red} illustrates that tokens with similar semantics in the Llama-3-8B-Instruct \citep{dubey2024llama} decoder model cluster coherently within the embedding space. For instance, synonyms of \emph{red} (e.g., \emph{ruby}, \emph{crimson}, \emph{vermilion}) form a compact cluster, while vehicle-related tokens (e.g., \emph{bus}, \emph{car}) occupy distinct region. In addition, words that are semantically ambiguous or participate in multiple contexts exhibit overlapping neighborhoods, suggesting the presence of multiple latent senses. This structured organization of the embedding space motivates the construction of a sense dictionary, where clustered representations can serve as discrete sense prototypes for downstream modeling.

However, directly applying SKD to decoders is challenging. In encoder models, the hidden state at position $t$ corresponds directly to the input token, so the model can utilize the sense dictionary for the token at position $t$ in the final hidden layer. In contrast, decoder models can generate any token from the entire vocabulary, so it is impractical to use the whole sense dictionary during inference for each position. To address this limitation, we use the sense dictionary as external knowledge during training instead of replacement for contextual embeddings. Thus, the sense dictionary serves solely as an external resource to support the training of the student model, which is fully self-contained -- given an input corpus, it can independently generate text in the same manner as a standard language model. In addition, inspired by HG2Vec \citep{wang2022hg2vec}, which leverages thesauri to capture lexical relations, we employ synonym and antonym relationships to strengthen the training framework.

We introduce the Decoder-based Sense Knowledge Distillation (DSKD) method that bridges word-level lexical knowledge with token blevel training for decoder-style large language models. Unlike prior approaches, DSKD allows generative models to inherit structured semantics from linguistic resources without requiring any additional steps during inference. We demonstrate that DSKD improves the performance of knowledge distillation for decoders and integrates the structured lexical knowledge into large-scale generative models (e.g., it can cut down the training cost by a factor of 2, and yet retain competitive performance). Our main contributions are as follows:
\begin{itemize}
\item We propose a new sense-based distillation paradigm for generative or decoder models, where the student model learns from the teacher's word senses.
    \item We enhance the sense distillation with root-level lexical information and word-level relationships like synonyms and antonyms as part of the knowledge distillation process for decoder models, leading to improved performance.
    \item We demonstrate that structured lexical semantics can be injected into decoder-based generative models only during training, without introducing any additional inference-time cost or architectural constraints.
\end{itemize}

\section{Related Works}
\paragraph{Decoder-only language models:}
Modern LLMs are predominantly \emph{decoder-only} architectures that rely on autoregressive next-token prediction with causal self-attention \citep{naveed2025comprehensive,matarazzo2025survey}. The GPT series pioneered this paradigm, and GPT-3 \citep{brown2020language} demonstrates that large-scale pretraining with transformer decoders achieves strong generalization across tasks. Building on this foundation, Llama 3  \citep{dubey2024llama} enhances downstream performance and long-context reasoning through efficient rotary position encodings. The Mistral models \citep{jiang2023mistral7b} introduce grouped-query attention and sliding-window attention to extend context handling. We use these two as the base LLM models to show the effectiveness of our decoder sense-based distillation.

\paragraph{Sense Embeddings:}
In contrast to single-vector word embeddings, sense embeddings associate multiple vectors with each word to capture its distinct meanings. LMMS  \citep{loureiro2022lmms} derives sense representations using annotated data. AutoExtend \citep{rothe2015autoextend} focuses on learning embeddings for synsets (a group of synonyms) and lexemes (associating spellings with particular meanings), treating both words and synsets as a sum of their lexemes.  Recent studies show that token embeddings remain strong carriers of semantic information \citep{mahajan2024revisiting,han2023word}, but these works lack evaluation on standard NLP benchmarks. To address the integration of sense representations into large-scale models, SKD \citep{wang2025multi} constructs a sense dictionary by clustering contextual token embeddings and performs sense-aware knowledge distillation, but it requires replacing the contextual embeddings with sense embeddings during inference. In contrast, our DSKD method enhances the sense dictionary with lexical resources such as dictionaries and thesauri to guide the training of a standalone sense-based student model that does not require sense embeddings or replacement during inference. 

\paragraph{Knowledge distillation:}
Knowledge distillation (KD) is a widely used model compression strategy in which a smaller student model is trained to approximate the behavior of a larger teacher model. Thus, KD ideally reduces model size while maintaining competitive performance. Early methods primarily targeted output-layer distribution matching, such as DistilBERT \cite{Sanh2019} and LIGHTPAFF \cite{song2020lightpaff}. For decoder-based LLMs, recent KD research proposes novel strategies that redesign training objectives, teacher signals, and training curricula. GKD \citep{agarwal2024policy} addresses the train–inference distribution gap by training students on their own generated sequences while using teacher feedback for correction. MCKD \citep{zhao-etal-2024-multistage} incorporates cross-partition labeling and multi-stage distillation to improve performance in low-resource semi-supervised sequence generation. Similarly, MD \citep{chenglin-etal-2024-mixed} enhances the multi-step reasoning of small language models by introducing well-balanced mixed-thought data from the teacher model, and DDK \citep{liu2024ddk} stabilizes and strengthens the distillation process by dynamically adapting the dataset composition according to domain-specific performance differences between the teacher and student models. In contrast, our DSKD method proposes a novel sense-based distillation approach that incorporates word senses along with both synonym and antonym relationships to perform distillation.

\section{DSKD Sense Dictionary}

\subsection{Token-level Sense Dictionary Construction}

We construct a token-level sense dictionary from a pretrained teacher model using the English Wikipedia dump \citep{wikipedia_dump_20240320} as the corpus source along with the training datasets mentioned in Section 5.2. We perform a single forward pass over the corpus and collect contextual embeddings for each token occurrence from the teacher’s last hidden layer, and we retain at most 2000 contextual embeddings per token. For each token $t$ in the vocabulary, we cluster its collected contextual embeddings using $k$-means \citep{macqueen1967some}. The resulting cluster centroids are treated as \emph{sense embeddings} of token $t$, each representing a prototypical semantic usage of the token under different contexts. Formally, let $k$ denote the number of clusters and $d$ denote the hidden dimension of the teacher model, so the sense embeddings of token $t$ are defined as
\[
S_t = [ s_{t,1}; s_{t,2}; \ldots; s_{t,k} ] \in \mathbb{R}^{k \times d}.
\]
where each $s_{t,i}$ corresponds to a centroid produced by $k$-means clustering. This dictionary provides a discrete, compact representation of token-level semantic variability and serves as the foundational semantic resource for the remainder of our framework.
\subsection{Lexical Relationship Construction}

While token-level sense embeddings capture semantic regularities from contextual usage, they do not explicitly encode structured lexical knowledge such as synonymy and antonymy. Such relationships provide complementary semantic signals: synonyms indicate semantic similarity beyond surface co-occurrence, while antonyms explicitly encode semantic contrast. Incorporating both types of relations allows us to constrain the semantic space with positive and negative lexical evidence, which is particularly beneficial for supervising representation learning.

To construct lexical relationships, we follow the pipeline introduced in HG2Vec~\cite{wang2022hg2vec} and extract synonym and antonym word pairs from two lexical resources: Wiktionary, accessed via Wiktextract~\cite{ylonen2022wiktextract}, and Roget’s Thesaurus~\cite{mccutcheon_2010}. Each extracted pair is treated as a lexical relationship edge between two words and is labeled as either a synonym or an antonym relation.

In addition to existing synonym and antonym pairs, we exploit morphological negation as a systematic and productive source of lexical contrast. In English, many oppositional meanings are formed through negating prefixes and suffixes such as un-, in-, dis-, and -less. Leveraging these regular morphological patterns allows us to substantially expand antonym relations, which are otherwise much sparser than synonym relations in standard lexical resources.

To incorporate morphological negation in a principled manner, we introduce MorphoLEX-English~\citep{sanchez2018morpholex}, which provides morphological decompositions and base forms. For each word $w$ that contains a negative prefix or suffix, we recover its corresponding base form $\mathrm{base}(w)$. Suppose a lexical relation is defined between two words $w_1$ and $w_2$, and $w_1$ contains a negating prefix or suffix. If $w_1$ and $w_2$ are labeled as synonyms, then the relation between $\mathrm{base}(w_1)$ and $w_2$ is treated as an antonym relation; conversely, if $w_1$ and $w_2$ are labeled as antonyms, then the relation between $\mathrm{base}(w_1)$ and $w_2$ is treated as a synonym relation.  Table~\ref{tab:morph_relation_examples} presents representative examples illustrating how morphological negation systematically transforms lexical relations.

Importantly, all lexical relationships are defined entirely at the word level and prior to tokenization. This design avoids the dependence on a specific tokenizer and lets these relations to be applied consistently in later stages of the framework.


\begin{table*}[t]
\centering

\begin{tabular}{l l | l l}
\hline
Original Pair & Original Relationship & New Pair & New Relationship \\
\hline
quit, discontinue     & synonym  & quit, continue        & antonym \\
accurate, faultless  & synonym  & accurate, fault     & antonym \\
opposite, dissimilar   & synonym  & opposite, similar    & antonym \\
\hline
variable, unchangeable & antonym & variable, changeable & synonym \\
unkind, friendly     & antonym  & kind, friendly      & synonym \\
disinterest, zeal    & antonym  & interest, zeal      & synonym \\
\hline
\end{tabular}
\caption{Examples of lexical relation transformation under morphological negation.}
\label{tab:morph_relation_examples}
\end{table*}

\subsection{Word-level Sense Composition}

In the previous sections, we introduced token-level sense embeddings (Section~3.1) and word-level lexical relations (Section~3.2). In practice, however, a word may correspond to either a single token or multiple tokens depending on the tokenizer. When a word is represented by a single token, its sense embeddings are directly given by the token-level sense dictionary. For words that are decomposed into multiple tokens by the tokenizer, we construct word-level sense embeddings by composing the sense embeddings of their constituent tokens.

Let a word $w$ be tokenized into a sequence of tokens $(t_1, t_2, \ldots, t_m)$, where $m$ denotes the number of tokens associated with the word. Recall that each token can have $k$ senses. If $m=1$, the word-level sense embeddings are given directly by the sense embeddings of the only token $S_{t_1}$. Otherwise, we compose sense embeddings iteratively across the token sequence.  We begin with the sense embeddings of the first token, $S_{t_1}$. At each subsequent step $h = 2, \ldots, m$, we first compute the mean of the composed sense embeddings obtained from tokens $(t_1,\ldots,t_{h-1})$. Using this mean representation, we perform nearest-neighbor matching under L2 distance with the sense embeddings $S_{t_h}$ of the next token, which can have up to $k$ senses. The matched sense embeddings are then incorporated into the composed representation.  After processing all tokens, the final word-level sense embeddings are obtained by averaging the aligned sense embeddings across the $m$ tokens: \[ \tilde{S}_w = \frac{1}{m} \sum_{h=1}^{m} \text{aligned}(S_{t_h}) . \]  This formulation ensures that each token contributes equally to the final word-level representation, avoids exponential attenuation of early tokens, and allows later tokens to iteratively refine sense alignment through nearest-neighbor matching. The resulting word-level sense embeddings provide a coherent semantic representation for words composed of multiple tokens. While this procedure describes the construction of a single
composite sense, we apply this composition independently for each of
the $k$ token-level senses associated with the word. As a result, each word is represented up to $k$ sense embeddings. Furthermore, we only decompose the word occurs in synonym and antonym pairs and incorporate them into our sense dictionary.

 To bound the complexity of word-level composition during dictionary construction, we limit the maximum number of tokens per word to $m$. Table~\ref{tab:m_sensitivity} reports the keep rates of synonym and antonym relations under different values of $m$, where the keep rate denotes the percentage of tokenized lexical relations whose word token length does not exceed $m$. As shown in the table, setting $m=3$ already retains over 90\% of synonym antonym relations for both tokenizers. Increasing $m$ beyond this point yields only marginal improvements in coverage while substantially increasing computational cost. Based on this analysis, we set $m=3$ in all experiments.

\begin{table}[t]
\centering
\small
\begin{tabular}{c|cc|cc}
\hline
 & \multicolumn{2}{c|}{LLaMA} & \multicolumn{2}{c}{Mistral} \\
$m$ & Synonym (\%) & Antonym (\%) & Synonym (\%) & Antonym (\%) \\
\hline
1  & 48.11 & 60.09 & 34.97 & 41.48 \\
2  & 82.22 & 87.03 & 72.39 & 74.38 \\
3  & 95.06 & 97.86 & 91.21 & 93.96 \\
4  & 98.54 & 99.61 & 96.96 & 98.99 \\
5  & 99.42 & 99.91 & 98.65 & 99.68 \\
\hline
\end{tabular}
\caption{Sensitivity analysis of the maximum span length $m$.
Keep rate denotes the percentage of tokenized lexical relations whose span length does not exceed $m$.}
\label{tab:m_sensitivity}
\end{table}

\section{DSKD Methodology}

\begin{figure*}[th]
\centering
\includegraphics[width=0.9\textwidth]{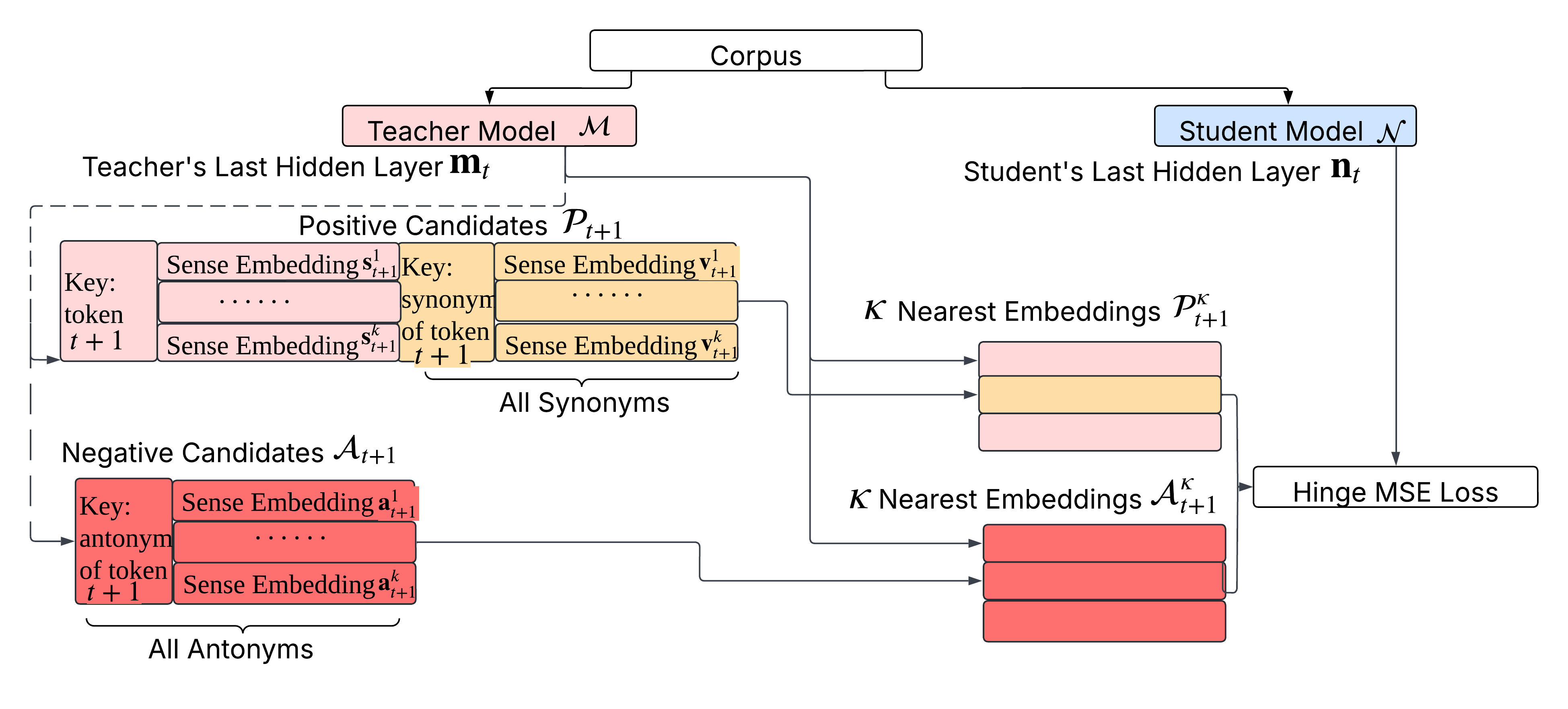} 
\caption{Overview of DSKD training: Given an input corpus, the teacher's last hidden layer $\mathbf{m}_t$ is used to retrieve sense embeddings of the token $t+1$ and its synonyms (positive candidates) and antonyms (negative candidates) from the sense dictionary. The $\kappa$ nearest positive and negative embeddings to $\mathbf{m}_t$ are selected, and the student's last hidden layer $\mathbf{n}_t$ is optimized with a hinge MSE loss. This process aligns the student’s semantic space with the teacher's sense dictionary.}
\label{fig:knowledge_distillation}
\end{figure*}

\subsection{Conceptual Overview}
\paragraph{Standard autoregressive language modeling.}
Decoder-only language models are trained with an autoregressive next-token prediction objective. Given a token sequence $\mathbf{x} = (x_1, \ldots, x_T)$, training minimizes the cross-entropy loss at each position,
\[
\mathcal{L}_{\mathrm{CE}}(t)
\;=\;
- \log p(x_{t+1} \mid x_{\le t}),
\]
which encourages the hidden representation at position $t$ to encode the contextual and semantic information for predicting the subsequent token $x_{t+1}$.

Although autoregressive models generate new tokens sequentially at inference time, during training this objective is evaluated at every position of the input sequence. As a result, a single forward pass computes hidden representations and token-level predictions for all tokens in the corpus, rather than only generating tokens after the input corpus.

However, the internal representations formed during training may not always align with the tokens actually presented to the user. For example, given the original corpus sentence, \textit{"She published her first book at 19, a collection of short fictitious ..."}, a LLaMA-3 decoder may internally generate alternative reconstructions, such as \textit{"her first book, the 19, and collection of poems stories ..."} in its hidden representations. They are not directly observable at the output level, because during training the model’s generated alternatives are overridden by the original corpus tokens. As a result, when training student models in knowledge distillation, relying solely on teacher hidden outputs may lead to suboptimal semantic alignment. Therefore, using the original corpus as an enhancement provides a more precise and stable semantic target.

\paragraph{Knowledge Distillation.}
Knowledge distillation trains a compact student model to mimic the behavior of a larger teacher model while preserving the standard autoregressive training structure.
At each position $t$, both the teacher and the student process the same input prefix $x_{\le t}$, and supervision is applied at the same alignment point through the prediction of the next token $x_{t+1}$.

Following prior work \citep{wang2025multi, Sanh2019}, the training objective at position $t$ is defined as
\[
\mathcal{L}_{KD}(t)
=
\mathcal{L}_{CE}(t)
+
\alpha \cdot \mathcal{L}_{KL}(t),
\]
where $\alpha > 0$ is a hyperparameter.
The cross-entropy term $\mathcal{L}_{CE}(t)$ trains the student model to predict the ground-truth next token $x_{t+1}$, consistent with standard language model training in which each hidden state $h_t$ is optimized to support next-token prediction \citep{dubey2024llama,jiang2023mistral7b}.
The distillation term $\mathcal{L}_{KL}(t)$ applies a temperature-scaled softmax to both student and teacher logits and minimizes the Kullback--Leibler divergence \citep{kullback1951information} between the two output distributions, encouraging the student to match the teacher’s predictive behavior. In practice, combining $\mathcal{L}_{CE}(t)$ and $\mathcal{L}_{KL}(t)$ provides more stable and effective supervision than using either term alone.

\paragraph{Decoder-based Sense Knowledge Distillation.}
DSKD strengthens the training process by augmenting standard knowledge distillation with sense-level supervision grounded in the input corpus. At each training position $t$, DSKD treats the ground-truth input token as the primary semantic anchor and explicitly incorporates its associated synonym and antonym senses as additional training signals.

Concretely, instead of relying solely on the teacher model’s output distribution, DSKD leverages a pre-constructed sense dictionary to identify a structured semantic neighborhood for the input token.
During training, the student is encouraged to encode the hidden representation at position $t$ in a way that is consistent with this sense neighborhood, thereby reinforcing semantic alignment with the original corpus token and its lexical relations. This supervision is applied at the same token position used for autoregressive next-token prediction and operates entirely within the training phase.
DSKD does not alter the autoregressive objective or the decoding procedure, but enhances representation learning by explicitly guiding how semantic information from the input corpus is organized in the student’s hidden space. Figure \ref{fig:knowledge_distillation} provides an overview of the DSKD training pipeline, illustrating how the teacher’s hidden representations are used to retrieve sense embeddings for the ground-truth token and its lexical neighbors, and how these signals guide the optimization of the student model as described next.

\subsection{Training Progress}
\paragraph{Sense dictionary lookup} As described earlier, conventional language modeling optimizes the next-token objective $\mathcal{L}_{CE}$.
We extend this training paradigm by augmenting supervision with lexical relations (extracted as described in Section 3.2) anchored to the ground-truth input tokens.
Specifically, for each token in the input corpus, we incorporate its associated synonym and antonym information from a pre-constructed sense dictionary as additional training signals. By grounding supervision in the input tokens and their lexical neighborhoods, we encourage the learned representations to capture finer-grained semantic structure beyond exact token-level prediction.

We randomly sample token positions $t$ for supervision and retrieve the sense embeddings of the next-token target $x_{t+1}$ from the sense dictionary whenever available, denoted as $\mathbf{s}_{t+1}$. In our corpus, approximately 14\% of tokens have associated synonyms and antonyms. For sampled tokens that have associated synonym or antonym relations, we additionally retrieve the sense embeddings of their synonyms $\mathbf{y}_{t+1}$ and antonyms $\mathbf{a}_{t+1}$; otherwise, we set $\mathbf{y}_{t+1}=\varnothing$ and $\mathbf{a}_{t+1}=\varnothing$. 
The positive and negative candidate sets are then defined as
\[
    \mathcal{P}_{t+1} = \mathbf{s}_{t+1} \cup \mathbf{y}_{t+1}
    \qquad 
    \mathcal{A}_{t+1} = \mathbf{a}_{t+1}
\]

\paragraph{Semantic consistency loss}


We introduce semantic consistency loss to encourage the student model’s last hidden state $\mathbf{n}_t$ to stay close to the sense embedding and synonyms of token $t+1$, while being pushed away from its antonyms.
However, a single token may correspond to multiple different senses, and thus its synonyms may also convey diverse meanings. Consequently, not all vectors in the positive set $\mathcal{P}_{t+1}$ represent the intended meaning to be predicted. To address this, we use the embedding of the teacher model’s last hidden layer, denoted as $\mathbf{m}_t$, as guidance. We select the $\kappa$ nearest neighbors (we use $\kappa=5$ in our experiments) to $\mathbf{m}_t$ from $\mathcal{P}_{t+1}$ as the positive candidates $\mathcal{P}_{t+1}^{\kappa}$.
Likewise, for antonyms, we also select the $\kappa$ nearest neighbors to $\mathbf{m}_t$ from $\mathcal{A}_{t+1}$ as the negative set $\mathcal{A}_{t+1}^{\kappa}$. Finally, we obtain:

\begin{align*}
\mathcal{P}_{t+1}^{\kappa} &= \operatorname{Top\kappa}_{\mathbf{p} \in \mathcal{P}_{t+1}}
\big(\| \mathbf{m}_t - \mathbf{p} \|_2\big) \\
\mathcal{A}_{t+1}^{\kappa} &= \operatorname{Top\kappa}_{\mathbf{a} \in \mathcal{A}_{t+1}}
\big(\| \mathbf{m}_t - \mathbf{a} \|_2\big)
\end{align*}

For a token at position $t$, we minimize the mean-squared error (MSE) between the student representation $\mathbf{n}_t$ and the selected positive candidates 
$\mathcal{P}_{t+1}^{\kappa}$, while enforcing a margin against the selected 
negative candidates $\mathcal{A}_{t+1}^{\kappa}$. In particular, 
let $d$ be the embedding dimension and define

\[
    \operatorname{MSE}(\mathbf{x}, \mathbf{y})
    = \tfrac{1}{d}\,\lVert \mathbf{x} - \mathbf{y} \rVert_2^2
\]

With hyperparameters $\beta_p>0$, $\beta_n>0$, and margin $\gamma>0$, 
the loss is then expressed as

\begin{align*}
&\mathcal{L}_{\text{sem}}(t) =
   \beta_p \;
   \frac{1}{\lvert \mathcal{P}_{t+1}^{\kappa}\rvert}
   \sum_{\mathbf{p} \in \mathcal{P}_{t+1}^{\kappa}} 
   \operatorname{MSE}(\mathbf{n}_t,\mathbf{p}) \notag \\
&\quad+\;
   \beta_n \;
   \frac{1}{\lvert \mathcal{A}_{t+1}^{\kappa}\rvert}
   \sum_{\mathbf{a} \in \mathcal{A}_{t+1}^{\kappa}}
   \big[\gamma-\operatorname{MSE}(\mathbf{n}_t,\mathbf{a})\big]_+
\end{align*}

where $[x]_+ = \max(x,0)$ denotes the hinge function. 
When both $\lvert \mathcal{P}_{t+1}^{\kappa}\rvert$ and $\lvert \mathcal{A}_{t+1}^{\kappa}\rvert$ are equal to $\kappa$, this formulation reduces to a standard hinge loss. When the cardinalities vary (e.g., some tokens have fewer than $\kappa$ synonyms or antonyms), averaging within each set ensures that positive and negative contributions are weighted equally in the objective, regardless of candidate count.

Finally, we define the overall objective as the sum of the knowledge distillation loss and the semantic consistency loss:
\[
\mathcal{L}_{\text{DSKD}}(t) = \mathcal{L}_{KD}(t) + \mathcal{L}_{sem}(t)
\]

By default, we train only the final two decoder layers of the student model to reduce training time while preserving the representational knowledge inherited from the teacher model. We further examine the effectiveness of this design in our ablation study.

\subsection{Inference}

Once the student model is trained, it functions as a standalone distilled model with its own input embedding layer, decoders layers, and output layers. Given an input corpus, it generates text using the standard language model. Because the student has been trained to align with the teacher’s outputs, it no longer depends on the sense dictionary at inference. 
Note that this is unlike SKD \citep{wang2025multi}, which requires replacing contextual embeddings with vectors from the sense dictionary during inference.
\section{Experiments}

\subsection{Experiment Setup}

We choose Llama-3-8B-Instruct \citep{dubey2024llama} and Mistral-7B-Instruct-v0.1 \citep{jiang2023mistral7b} as the teacher decoder models; both have 32 layers. 
We compare DSKD with a standard KD approach that simply uses the reduced number of student layers to minimize the $\mathcal{L}_{KD}$ loss.
In general, increasing the number of layers $L$ for the student model leads to improved performance. When all $32$ layers are used, performance reaches the same level as the teacher; however, for this setting, no compression is achieved. At the other extreme, using fewer than five layers results in very poor performance (below $30\%$ accuracy), and thus these are not practically useful. Therefore, we set the number of layers to 16 so the baseline KD model achieves approximately $70$--$90\%$ of the teacher’s F1-score or accuracy. We then augment with DSKD loss following the same setting.

To reduce training time and memory usage, we copy and freeze the embedding layer, the first 14 decoder layers, and the output layer, while training only the last 2 layers. The impact of the number of trainable layers is analyzed in our ablation study. 
We conducted all experiments on 6 NVIDIA V100 GPUs (32\,GB each) paired with 20-core IBM Power9 processors and 512\,GB of RAM, with training completed within 12 hours. We use bf16 percision for all the experiments. Finally, note that for all the benchmark datasets (except MMLU~\citep{hendrycks2020measuring}) we show inference results on the validation set, which is not used for training, since their test splits are not public (requiring server submission) with hyperparameter selected using the subset of training set.

\subsection{Dataset Setup}
In terms of benchmark datasets, we choose ARC \citep{clark2018think}, which consists of multiple-choice science questions from standardized exams, requiring multi-step reasoning and background knowledge; CommonsenseQA \citep{talmor2018commonsenseqa}, which evaluates commonsense reasoning through five-choice questions with carefully constructed plausible distractors; MMLU \citep{hendrycks2020measuring}, covering 57 subjects across STEM, humanities, social sciences, and professional domains, and serving as a comprehensive benchmark for general-purpose reasoning and knowledge in large language models; and PIQA \citep{bisk2020piqa}, which focuses on physical commonsense by asking models to choose the more plausible of two solutions to everyday tasks.

To support training on MMLU, which does not provide an official training set, we leverage the MMLU-Pro \citep{wang2024mmlu} training split and convert the original six-choice format to four choices, while filtering out any questions that overlap with the original MMLU dataset to prevent data leakage. Across ARC-Challenge, CommonsenseQA, MMLU-Pro, and PIQA, we use 25\% of the available training data for training and use grid search for hyperparameters. Model evaluation is conducted on the MMLU test set and the validation sets of ARC-Challenge, CommonsenseQA, and PIQA, all under a 5-shot setting.

In addition to classification benchmarks, we use SQuADv2~\citep{rajpurkar2018know} as a benchmark for generative tasks, as it offers a large-scale dataset containing both answerable and unanswerable questions. This setting requires models not only to generate correct answers when sufficient evidence is available but also to abstain appropriately when the passage does not support an answer. For SQuADv2, we train the models and perform grid search using 25\% of the training split, and report F1 scores on the full validation set. All experiments on SQuADv2 are conducted in a 1-shot setting, where the demonstration example is sampled from a different question within the same paragraph.

\subsection{Performance}
\begin{table*}[t]
\resizebox{\textwidth}{!}{%
\begin{tabular}{|l | c | c | c | c | c | c|}
\hline
Model & \# of Layers & ARC & CSQA & MMLU & PIQA & SQuAD \\
\hline
LLaMA3-8B & 32 & 81.45$\pm$1.07 & 78.08$\pm$0.41 & 67.22$\pm$0.20 & 78.49$\pm$1.33 & 72.90$\pm$0.13 \\
KD        & 16 & 77.89$\pm$0.42 & 76.43$\pm$0.45 & 61.26$\pm$0.59 & 81.78$\pm$0.34 & 68.58$\pm$0.13 \\
DSKD      & 16 & \textbf{79.66$\pm$0.34} & \textbf{76.70$\pm$0.17} & \textbf{64.38$\pm$0.38} & \textbf{82.63$\pm$0.27} & \textbf{69.11$\pm$0.36} \\
\hline
Mistral-7B & 32 & 70.33$\pm$0.57 & 68.34$\pm$0.75 & 55.37$\pm$0.27 & 73.72$\pm$0.59 & 66.26$\pm$0.13 \\
KD         & 16 & 61.11±0.31 & 66.16$\pm$0.78 & 52.04$\pm$0.43 & 78.88$\pm$0.22 & 57.82$\pm$0.25 \\
DSKD       & 16 & \textbf{64.78$\pm$0.43} & \textbf{68.44$\pm$0.70} & \textbf{54.44$\pm$0.37} & \textbf{79.53$\pm$0.45} & \textbf{58.09$\pm$0.18} \\
\hline
\end{tabular}
}
\caption{Performance comparison between student models trained with KD and DSKD. Bold numbers indicate the best results among the student models.}
\label{tab:dskd_results}
\end{table*}

Table~\ref{tab:dskd_results} reports the performance of the teacher models and their distilled counterparts on ARC, CSQA, MMLU, PIQA, and SQuAD. Overall, DSKD consistently outperforms standard KD across both LLaMA-3-8B-Instruct and Mistral-7B-Instruct backbones, demonstrating robust gains across a diverse set of reasoning, knowledge-intensive, and reading comprehension benchmarks. The improvements are particularly evident on tasks that require broader knowledge integration and semantic disambiguation.

For LLaMA-based models, DSKD yields consistent improvements over KD on all evaluated benchmarks. On ARC and MMLU, DSKD improves accuracy from $77.89\%$ to $79.66\%$ and from $61.26\%$ to $64.38\%$, respectively, highlighting its effectiveness on reasoning-heavy and knowledge-intensive tasks where semantic ambiguity plays a critical role. On CSQA, where the performance gap between the teacher and student models is already narrow, DSKD still achieves a modest but stable improvement over KD, indicating that sense-aware supervision remains beneficial even near performance saturation. On PIQA, KD already provides a strong boost over the teacher, and DSKD further improves performance to $82.63\%$ while maintaining low variance. On SQuAD, DSKD consistently outperforms KD, increasing accuracy from $68.58\%$ to $69.11\%$, although the relative gain is smaller than that observed on ARC and MMLU due to the increased difficulty of precise span-level prediction. On SQuAD, DSKD again outperforms KD, increasing accuracy from $68.58\%$ to $69.11\%$, but the improvement remains more modest, reflecting the higher difficulty and tighter supervision requirements of extractive question answering.

A similar but more pronounced trend is observed for Mistral-based models. Standard KD suffers from substantial degradation relative to the teacher, for example reducing ARC accuracy from $70.33\%$ to $61.11\%$ and MMLU from $55.37\%$ to $52.04\%$. In contrast, DSKD recovers a significant portion of this performance gap, improving ARC to $64.78\%$ and MMLU to $54.44\%$. Consistent gains are also observed on CSQA and PIQA, where DSKD improves accuracy from $66.16\%$ to $68.44\%$ and from $78.88\%$ to $79.53\%$, respectively.

We attribute this difference to the interaction between vocabulary size and semantic granularity. LLaMA employs a substantially larger vocabulary, which implicitly distributes semantic variation across more tokens and makes standard KD relatively effective. By contrast, Mistral relies on a smaller vocabulary, forcing individual tokens to encode a wider range of senses, which exacerbates semantic compression during distillation. DSKD alleviates this issue by explicitly introducing sense-level supervision, enabling the student model to better preserve semantic distinctions under tighter vocabulary constraints. In summary, these results demonstrate that DSKD provides robust improvements over standard KD. 

\section{Ablation Studies}

\begin{figure}[ht]
\centering
\includegraphics[width=0.475\textwidth, height=1.5in]{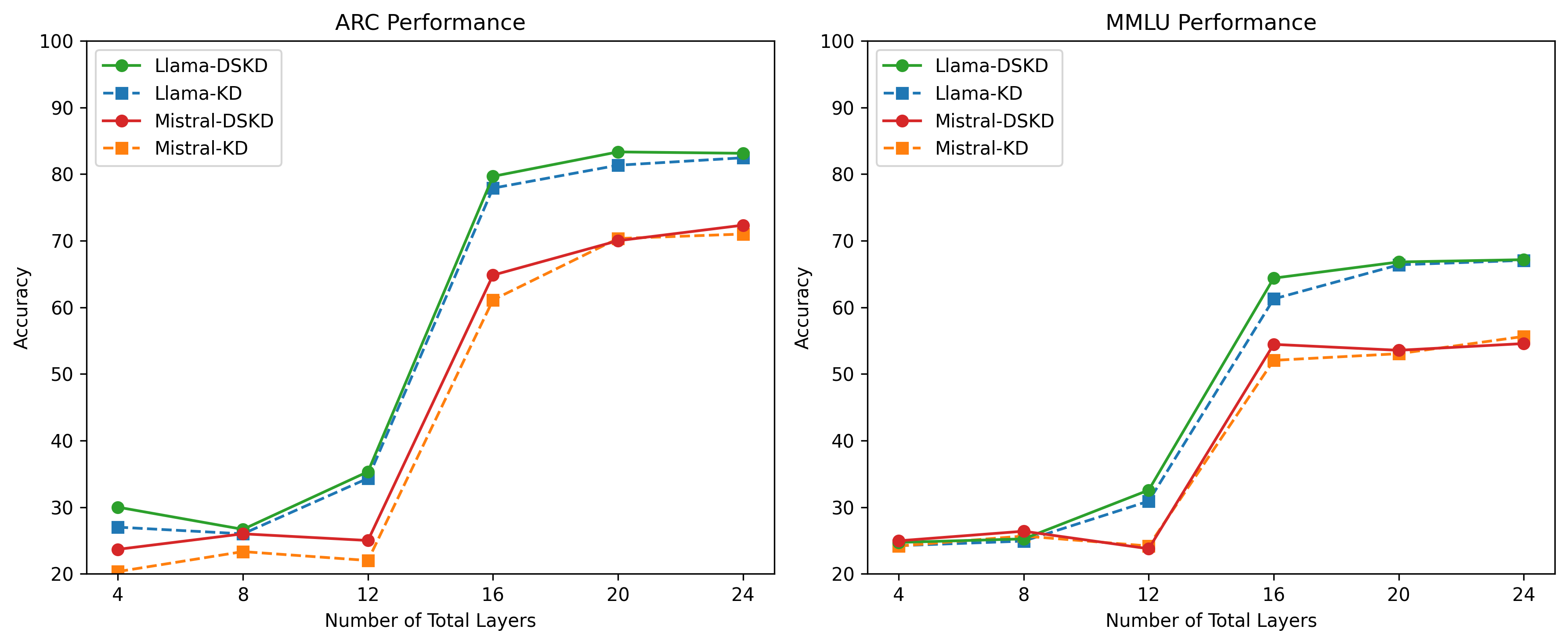}\\
(a)
\includegraphics[width=0.475\textwidth, height=1.5in]{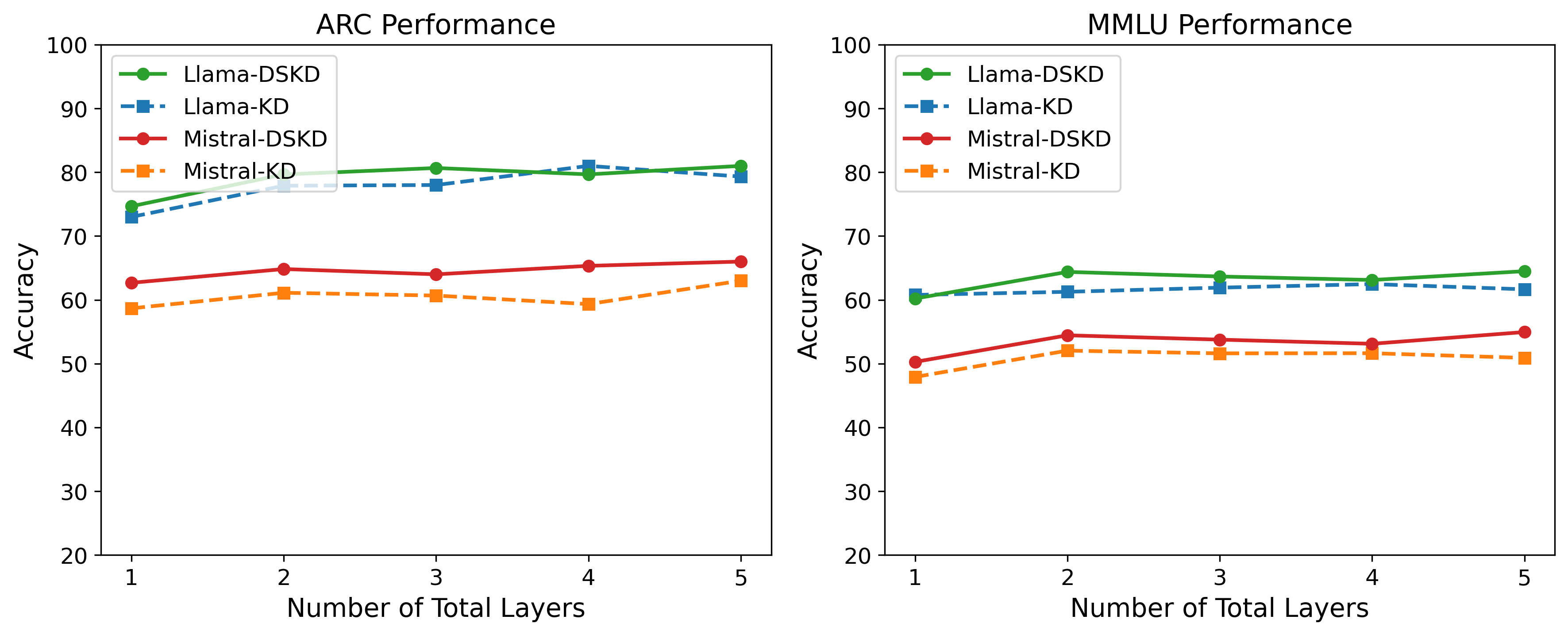}\\
(b)
\caption{Performance of KD and DSKD distilled from Llama-3-8b-Instruct on ARC and MMLU across varying (a) 
total number of layers, and (b) number of trainable student layers.
}
\label{fig:ab_total_layer}
\label{fig:ab_trainable_layer}
\end{figure}

\subsection{Number of Layers}
\paragraph{Number of total student layers}
To examine the impact of different total number of student layers on the performance of the student models, we conduct an ablation study on ARC and MMLU. Figure~\ref{fig:ab_total_layer} (a) reports the results across different student depths for both KD and DSKD. For both KD and DSKD, performance exhibits a small drop at layer 16 on both datasets, followed by sharp declines if the number of layers is lowered beyond these points. Based on this trend, we select layer layer 16 in our experiments. Moreover, DSKD outperforms KD across most of the configurations.

\paragraph{Number of training student layers}
In contrast to the impact of the total number of student layers, varying the number of trainable student layers leads to relatively minor performance changes. Figure~\ref{fig:ab_trainable_layer} (b) presents ARC and MMLU results with the total student layers fixed at 16, while the number of trainable layers ranges from 1 to 5 for both KD and DSKD. Compared to using a single trainable layer, employing 2 trainable layers consistently yields better performance across tasks. However, further increasing the number of trainable layers does not lead to sustained improvements, while incurring additional training cost. Accordingly, 2 trainable layers are used in all subsequent experiments.

\subsection{Sense Selections}
\begin{table}[t]
\centering
\begin{tabular}{|c|cc|cc|}
\hline
\textbf{$k$} & \multicolumn{2}{c|}{\textbf{LLaMA}} & \multicolumn{2}{c|}{\textbf{Mistral}} \\
\cline{2-5}
 & \textbf{ARC} & \textbf{MMLU} & \textbf{ARC} & \textbf{MMLU} \\
\hline
5  & 77.33 & 62.18 & \textbf{64.78} & \textbf{54.44} \\
10 & \textbf{79.66} & \textbf{64.38} & 64.33 & 54.25 \\
15 & 79.00 & 64.57 & 65.00 & 54.91 \\
20 & 81.33 & 64.68 & 64.00 & 53.68 \\
25 & 80.00 & 65.23 & 65.33 & 54.99 \\
\hline
\end{tabular}
\caption{Effect of the number of clusters $k$ on ARC and MMLU benchmarks. The selected value of $k$ used in subsequent experiments is highlighted in bold.}
\label{tab:k_ablation}
\end{table}

\paragraph{Number of clusters $k$} Table \ref{tab:k_ablation} presents the effect of varying the number of clusters $k$ for both LLaMA and Mistral backbones. Overall, increasing $k$ generally leads to improved performance, indicating that a finer-grained sense partition can better capture semantic variability. However, the gains become marginal as $k$ grows, and no consistent or significant improvements are observed beyond moderate values. Moreover, larger $k$ substantially increases the size of the resulting sense dictionary, leading to higher memory and storage overhead. Taking this trade-off into account, we select $k=10$ for LLaMA-based models and $k=5$ for Mistral-based models in all subsequent experiments.

\begin{table}[t]
\centering
\begin{tabular}{|c|cc|cc|}
\hline
\textbf{$\kappa$} & \multicolumn{2}{c|}{\textbf{LLaMA}} & \multicolumn{2}{c|}{\textbf{Mistral}} \\
\cline{2-5}
 & \textbf{ARC} & \textbf{MMLU} & \textbf{ARC} & \textbf{MMLU} \\
\hline
1  & 78.00 & 63.93 & 63.67 & 53.60 \\
5  & \textbf{79.66} & \textbf{64.38} & \textbf{64.78} & \textbf{54.44} \\
10 & 79.33 & 64.20 & 64.33 & 54.35 \\
15 & 80.33 & 64.76 & 65.67 & 54.53 \\
20 & 79.00 & 64.71 & 66.00 & 53.68 \\
\hline
\end{tabular}
\caption{Effect of $\kappa$ on ARC and MMLU benchmarks.} 
\label{tab:kappa_ablation}
\end{table}

\paragraph{Number of selected senses during training $\kappa$} Table~\ref{tab:kappa_ablation} shows the impact of varying $\kappa$ for semantic consistency. When $\kappa=1$, performance is consistently lower, indicating that an insufficient weight on the sense-level objective limits the effectiveness of DSKD. Increasing $\kappa$ generally improves performance, but larger values do not necessarily lead to further improvements. Based on this observation, we select $\kappa=5$ as a balanced setting that yields strong performance across tasks and backbones.

\subsection{Training Hyperparameters}
\paragraph{DSKD loss coefficient $\beta_p$ and $\beta_n$} 

Table \ref{tab:kappa_ablation} reports the performance under different values of $\beta_p$ and $\beta_n$, where we tie the two coefficients. Overall, the results indicate that the choice of $\beta$ plays a non-trivial role in balancing the contribution of sense-level supervision. When $\beta$ is set too small, the effect of DSKD becomes marginal and the performance approaches that of standard KD, suggesting insufficient utilization of sense information. In contrast, overly large values of $\beta$ do not yield further improvements and may even hurt performance, as the sense-level objective begins to dominate the training process. Based on this trade-off, we select $\beta=1.0$ for LLaMA-based models and $\beta=1.5$ for Mistral-based models, which consistently achieve strong performance.

\begin{table}[t]
\centering
\begin{tabular}{|c|cc|cc|}
\hline
\textbf{$\beta_p$, $\beta_n$}  & \multicolumn{2}{c|}{\textbf{LLaMA}} & \multicolumn{2}{c|}{\textbf{Mistral}} \\
\cline{2-5}
 & \textbf{ARC} & \textbf{MMLU} & \textbf{ARC} & \textbf{MMLU} \\
\hline
0.5 & 77.67 & 61.42 & 64.00 & 52.07 \\
1.0 & \textbf{79.66} & \textbf{64.38} & 63.33 & 53.51 \\
1.5 & 79.33 & 63.45 & \textbf{64.78} & \textbf{54.44} \\
2.0 & 78.33 & 61.45 & 61.33 & 50.57 \\
\hline
\end{tabular}
\caption{Performance on ARC and MMLU under different values of $\beta_p$ and $\beta_n$.}
\label{tab:beta_ablation}
\end{table}

\section{Time and Space}

\begin{table}[!ht]
\centering
\resizebox{\columnwidth}{!}{
\begin{tabular}{|l|c|c|c|c|c|}
\hline
\textbf{Model} & \textbf{\# Layers} & \textbf{\# Params} & \textbf{Memory} & \textbf{Train} & \textbf{Eval} \\ 
 &  &  & & \textbf{ Time} & \textbf{ Time} \\ \hline
Llama-3-8B & 32 & 8.0B & 15.0GB & N/A & 45min \\ \hline
KD & 16 & 4.5B & 8.5GB & 5h02min & 23min \\ \hline
DSKD & 16 & 4.5B & 8.5GB & 5h17min & 22min \\ \hline
\end{tabular}
}
\caption{Comparison of model size, memory, training time per epoch, and MMLU evaluation time for Llama-3-8B and its student models.}
\label{tab:efficiency_mmlu}
\end{table}

 Table~\ref{tab:efficiency_mmlu} presents a comparison of KD and DSKD using the Llama-3-8B-Instruct model on the MMLU dataset, in terms of parameter size, memory usage, training time on the whole classification task, and evaluation time on the full MMLU benchmark. The student model uses half of the teacher’s layers, reducing the total parameters from 8.0B to 4.5B and the memory footprint from 15.0GB to 8.5GB—approximately $56\%$ of the teacher model with bf16 percision. DSKD introduces only a minor training-time overhead of around $5\%$ compared to KD, while maintaining nearly identical evaluation time. This confirms that DSKD adds minimal computational cost, as its additional semantic loss operates only during training. 
\section{Conclusion and Future Works}
In this paper, we introduce our decoder-based sense knowledge distillation framework that integrates sense dictionaries and lexical relationship resources into the training. 
By eliminating the reliance on dictionary lookups during inference, DSKD enables generative models to inherit structured lexical semantics while maintaining efficiency and broad applicability. Our experiments with Llama-3-8B-Instruct and Mistral-7B-Instruct demonstrate consistent improvements over standard knowledge distillation, highlighting the effectiveness of sense-guided supervision. 

Our next step is to develop a mechanism that assigns each token to a different number of clusters. In the current framework, a fixed number of senses $k$ is used for all tokens, which may not reflect the varying degrees of semantic ambiguity across words. Some tokens correspond to highly polysemous words and would benefit from a richer sense inventory, while others exhibit little ambiguity and require only a small number of senses. As a result, we can obtain a more flexible and efficient representation that better reflects the intrinsic semantic complexity of individual tokens.

\bibliographystyle{ACM-Reference-Format}
\bibliography{sample-base}

\appendix
\clearpage
\appendix
\section*{Appendix}

\section{Open Source Code}
We will release our project, including the training code, sense dictionary, and model weights, as open source project on GitHub after the review period. The implementation is built with NumPy~\cite{harris2020array}, PyTorch~\cite{paszke2019pytorch}, and scikit-learn~\cite{scikit-learn} for data processing and model training, with Matplotlib~\cite{hunter2007matplotlib} used for visualization.
We use Llama-3-8B-Instruct \citep{dubey2024llama} and Mistral-7B-Instruct \citep{jiang2023mistral7b} as the teacher models. We obtained research licenses for both models from Hugging Face and downloaded them using the \texttt{transformers} library \citep{wolf2020transformers}. All evaluation benchmarks are accessed through the Hugging Face \texttt{datasets} library \citep{lhoest2021datasets}.

\section{Dataset}
\subsection{Dictionary and Thesauri}

We build the sense dictionary from the Wikipedia dump \citep{wikipedia_dump_20240320} and the training dataset. We construct the relationship dictionary using synonym and antonym pairs extracted from Wiktionary \citep{ylonen2022wiktextract} and Roget’s Thesaurus through HG2Vec \citep{mccutcheon_2010,wang2022hg2vec}. To incorporate morphological information, we also integrate semantic root forms obtained from MorphoLEX-English \citep{sanchez2018morpholex}. Table~\ref{tab:dict_stats} summarizes the statistics of the dictionaries used in our experiments. 

\begin{table}[!ht]
\centering
\resizebox{\columnwidth}{!}{
\begin{tabular}{|l|c|c|}
\hline
\textbf{} & \textbf{Llama-3-8B-Instruct} & \textbf{Mistral-7B-Instruct} \\ \hline
Tokenizer vocab size & 128,256 & 32,000 \\ \hline
\# of Sense token keys & 69,832 & 27,541 \\ \hline
\# of Word token keys & 15,695 & 27,702 \\ \hline
Sense dict size & 7.08 GB & 2.06 GB \\ \hline
\# of synonym pairs & 923,950 & 923,950 \\ \hline
\# of antonym pairs & 75,949 & 75,949 \\ \hline
\end{tabular}
}
\caption{Vocabulary and dictionary statistics for Llama-3-8B and Mistral-7B models.}
\label{tab:dict_stats}
\end{table}

\subsection{Evaluation Benchmark}
We select SQuADv2 \citep{rajpurkar2018know} as for the general purpose generative task. We further evaluate our models on a diverse set of multiple choice benchmark datasets: ARC \citep{clark2018think}, CommonsenseQA \citep{talmor2018commonsenseqa}, MMLU \citep{hendrycks2020measuring} and PIQA \citep{bisk2020piqa}, for token-constrained generative tasks.

\section{Model Hyparameters}
\subsection{KD Hyparameter Ablation Study}
\begin{table}[t]
\centering
\begin{tabular}{|c|cc|cc|}
\hline
\textbf{$\alpha$} & \multicolumn{2}{c|}{\textbf{LLaMA}} & \multicolumn{2}{c|}{\textbf{Mistral}} \\
\cline{2-5}
 & \textbf{ARC} & \textbf{MMLU} & \textbf{ARC} & \textbf{MMLU} \\
\hline
0   & 75.67 & 61.25 & 58.67 & 51.75 \\
0.5 & 78.33 & 62.03 & 59.33 & 49.27 \\
\textbf{1}   & \textbf{77.89} & \textbf{61.26} & \textbf{61.11} & \textbf{52.04} \\
2   & 80.33 & 64.76 & 60.00 & 51.28 \\
\hline
\end{tabular}
\caption{Effect of $\alpha$ on ARC and MMLU benchmarks across LLaMA and Mistral backbones on KD. The selected value of $\alpha$ used in subsequent experiments is highlighted in bold.}
\label{tab:alpha_ablation}
\end{table}
Table~\ref{tab:alpha_ablation} presents an ablation study on the hyperparameter $\alpha$, which is defined in the standard KD objective and controls the relative weighting of the distillation loss. Based on the observed trade-off between task performance and distillation effectiveness, we select $\alpha=1.0$ for each backbone. To ensure a fair comparison, we then keep the same $\alpha=1.0$ setting when conducting all DSKD experiments.

\subsection{DSKD Hyparameter Tables}

For hyperparameter search, we explore learning rates in $\{5\mathrm{e}{-6}$, $1\mathrm{e}{-5}$, $2\mathrm{e}{-5}$, $5\mathrm{e}{-5}$, $1\mathrm{e}{-4}$, $2\mathrm{e}{-4}\}$. For the sampler sizes, we test values of $\kappa=\{1,\,5,\,10,\,15,\,20\}$. For the loss weights $\beta_p, \beta_n$, we test $\{0.5,\,1.0,\,1.5,\,2.0\}$. Table ~\ref{tab:parameter_dskd} presents the training hyperparameters used for KD and DSKD across all tasks with the Llama-3 and Mistral models.

\begin{table*}[!ht]
\centering
\begin{tabular}{|l|c|c|c|c|}
\hline
\textbf{} & \textbf{Llama-3 KD} & \textbf{Llama-3 DSKD} & \textbf{Mistral KD} & \textbf{Mistral DSKD} \\ \hline
Total layers ($L$) & 16 & 16 & 16 & 16 \\ \hline
Trainable layers & 2 & 2 & 2 & 2 \\ \hline
Learning rate & 0.00005 & 0.00005 & 0.00002 & 0.00002 \\ \hline
epoch & 2 & 2 & 2 & 2 \\ \hline
candidate sample size $\kappa$ & N/A & 5 & N/A & 5 \\ \hline
$\alpha$ & 1.0 & 1.0 & 1.0 & 1.0 \\ \hline
$\beta_p$ & N/A & 1.5 & N/A & 1.0 \\ \hline
$\beta_n$ & N/A & 1.5 & N/A & 1.0 \\ \hline
$\gamma$ & N/A & 1.0 & N/A & 1.0 \\ \hline
\end{tabular}
\caption{Full training hyperparameter configurations.}
\label{tab:parameter_dskd}
\end{table*}

\section{Model Comparison: Space/Parameters and Time}
Tables \ref{tab:training} and \ref{tab:time_comparison} provide additional comparisons of model efficiency in terms of parameter size, memory usage, training time, and evaluation latency. Table \ref{tab:training}summarizes the space and training-time cost of the teacher models and their corresponding student variants under KD and DSKD, showing that DSKD introduces only a marginal increase in training time while preserving the same memory footprint as KD. Table \ref{tab:time_comparison} further reports evaluation time across different benchmarks, where KD and DSKD exhibit nearly identical inference latency, indicating that the proposed sense-level supervision does not introduce additional runtime overhead during evaluation.
\begin{table*}[!ht]
\centering
\small
\begin{tabular}{|l|c|c|c|c|}
\hline
\textbf{Model} & \textbf{\# of Layers} & \textbf{\# of Parameters} & \textbf{Memory} & \textbf{Training Time}  \\ \hline
Llama-3-8B-Instruct & 32 & 8.0B & 15.0GB & N/A  \\ 
KD & 16 & 4.5B & 8.5GB & 5h02min  \\ 
DSKD & 16 & 4.5B & 8.5GB & 5h17min  \\   \hline
Mistral-7B-Instruct & 32 & 7.2B & 13.5GB & N/A \\ 
KD & 16 & 3.7B & 6.9GB & 4h53min  \\ 
DSKD & 16 & 3.7B & 6.9GB & 5h04min  \\ \hline
\end{tabular}
\caption{Comparison of space and training time.}
\label{tab:training}
\end{table*}

\begin{table*}[t]
\centering
\small
\begin{tabular}{|l|c|c|c|c|c|}
\hline
\textbf{Model}  & \textbf{ARC} & \textbf{CSQA} & \textbf{MMLU} & \textbf{PIQA} & \textbf{SQUADV2}\\
\hline
Llama-3-8B-Instruct   & 1min & 2.5min & 45min & 4.5min & 23min \\
KD                    & 1min & 1.5min & 23min & 3min  & 15min  \\
DSKD                  & 1min & 1.5min & 22min & 3min  & 15min   \\  \hline
Mistral-7B-Instruct   & 1min & 2.5min & 51min & 5min  & 21min   \\
KD                    & 1min & 1.5min & 25min & 3min  & 15.5min \\
DSKD                  & 1min & 1.5min & 26min & 3min  & 15min \\
\hline
\end{tabular}
\caption{Evaluation time comparison across models and tasks.}
\label{tab:time_comparison}
\end{table*}

\end{document}